\documentclass[a4paper,10pt]{article}

\usepackage[utf8]{inputenc}
\usepackage{geometry}
\usepackage{amsmath}
\usepackage{url}
\usepackage{mathptmx} 
\usepackage{caption}
\usepackage{array}
\usepackage{multirow} 
\usepackage{pdfpages}
\usepackage{tabularray}
\usepackage{fancyhdr}
\UseTblrLibrary{booktabs}

\usepackage{placeins}
\usepackage{adjustbox}
\usepackage[normalem]{ulem}

\usepackage{titling} 
\newgeometry{left=0.6in,right=0.6in,top=1in,bottom=1in}
\title{%
\rule{\textwidth}{1.5pt}
  \textbf{Deformable Attention Mechanisms Applied to Object Detection, case of Remote Sensing}
  \rule{\textwidth}{1.5pt}\\[1em]
  \large \textit{Paper accepted at the 29th International Conference on Knowledge-Based and Intelligent Information and Engineering Systems (KES 2025)}%
}

\author{Anasse Boutayeb \textsuperscript{*} \and Iyad Lahsen-cherif \textsuperscript{*} \and Ahmed El Khadimi \textsuperscript{*}}
\date{\today}
\fancypagestyle{firstpage}{ 
  \fancyhf{} 
  
  \fancyfoot[L]{\raisebox{-20pt}{\textsuperscript{*}  National Institute of Posts and Telecommunications (INPT)}} 
}

\begin{document}
\maketitle
\thispagestyle{firstpage}
\setlength{\columnsep}{12pt}
\noindent\textbf{Abstract-} Object detection has recently seen an interesting trend in terms of the most innovative research work, this task being of particular importance in the field of remote sensing, given the consistency of these images in terms of geographical coverage and the objects present. Furthermore, Deep Learning (DL) models, in particular those based on Transformers, are especially relevant for visual computing tasks in general, and target detection in particular. Thus, the present work proposes an application of Deformable-DETR model, a specific architecture using deformable attention mechanisms, on remote sensing images in two different modes, especially optical and Synthetic Aperture Radar (SAR). To achieve this objective, two datasets are used, one optical, which is Pleiades Aircraft dataset, and the other SAR, in particular SAR Ship Detection Dataset (SSDD). The results of a 10-fold stratified validation showed that the proposed model performed particularly well, obtaining an F1 score of 95.12\% for the optical dataset and 94.54\% for SSDD, while comparing these results with several models detections, especially those based on CNNs and transformers, as well as those specifically designed to detect different object classes in remote sensing images.
\\
\\
\\

\section{Introduction}
\label{main}
In addition to several Artificial Intelligence (AI) novel methods \cite{lahsen2021energy, lahsen2022real}, transformer-based models \cite{vaswani2017attention} represent a prominent trend in Deep Learning (DL), particularly with regard to highly complex tasks such as target detection. In particular, object detection on remote sensing images is a current challenge \cite{boutayeb2025machine}, given the increasing use of this mode by a large community of users. Moreover, this specific task, in both active and passive modes, represents a very delicate task, given the difficulty of recognizing objects, particularly small-scale ones, on the one hand, and the consistency of the concerned images, on the other. Consequently, extracting sufficient features in this context is an obstacle to be solved, notably by using the most innovative models. Given the wealth of information contained in remote sensing images, particularly for very high-resolution ones (less than 1m), target detection takes on greater importance. Several applications can take advantage of the products resulting from this process, such as agriculture \cite{ghosal2019weakly}, security \cite{ke2017military}, or even the field of energy and mining, where the automatic detection of solar panels \cite{malof2016automatic}, for example, represents a key requirement. 
\\
Indeed, the aim of this work is to present the application of a particular concept, especially deformable attention mechanisms, on two datasets. Therefore, the final purpose is to evaluate the performance of this model in terms of prediction accuracy and convergence speed, by comparing it with the models most widely used in the literature. In fact, the remainder of this paper will be organized as follows, Section 2 is devoted to the situation of this work in relation to similar research, while emphasizing its particular contribution. Then, Section 3 presents the models and datasets used for comparison, while Section 4 introduces the methodology by discussing the evaluation metrics and the benchmarking workflow. Section 5 illustrates the experimental results, mentioning the metric values, as well as the measurement of model speed based on training times, together with a rapid evaluation in order to compare Deformable-DETR with other transformer-based remote sensing detectors. Finally, the last section is devoted to the outlooks, illustrating the possible follow-up to this research as well as the main conclusions.
\section{Related works}
\label{main}
A multitude of recent research have addressed the issue of target detection on remotely sensed images. As an example, Coulson et al. \cite{coulson2025comparative} conducted a comparative study to present a set of CNN-based models. The authors compared several methods, namely Single Shor Detector (SSD) \cite{liu2016ssd}, YOLOv3, RetinaNet and Faster R-CNN, while obtaining the highest performance for the Faster R-CNN model. Besides, Aleissaee et al. \cite{aleissaee2023transformers} proposed the first survey of the application of Vision Transformers (ViT) \cite{dosovitskiy2020image} on remote sensing. With regard to object detection, the authors emphasized the difficulty of this task, due to scale variations, and the diversified orientations of the present objects. Also, the authors presented two main categories of models, specifically those based on detector transformers (DETR), and hybrid models, combining Convolutional Neural Networks (CNNs) with transformer-based architectures, including Arbitrary-Oriented Object DEtection TRansformer (AO2-DETR) \cite{dai2022ao2} and O2DETR \cite{ma2021oriented}.
\\
Given the interest and novelty of deformable attention mechanisms in computer vision tasks, particularly with regard to optimizing training and inference over computational costs, several previous researches have applied this principle to remote sensing field. Zhao et al. \cite{zhao2024road} propose DOCswin-Trans, an innovative method for road detection, based on the CSwin Transformer improved classifier and deformable attention in order to address the problem of scale variability. In another context, Ning et al. \cite{ning2025mabdt} present Multi-scale Attention Boosted Deformable Transformer (MADBT), a model for denoising remote sensing images. The authors used multiscale attention to extract fine local image features, and Attention Deformable Transformer Block (ADTB) to solve the problem of noise variations at different scales. 
\\
Furthermore, several research papers have addressed the subject of applying Deformable-DETR model to remote sensing images. For example, Chen et al. \cite{chen2024pr} present PR-Deformable DETR, a novel architecture applying adaptive pyramidal fusion on the base model, while achieving mAP values of 95.1\% and 88.3\% on RSOD and NWPU VHR-10 optical datasets. Plus, Li et al. \cite{li2025refined} propose Refined Deformable-DETR, which is a variant of Deformable-DETR, with innovative contributions including a multi-scale adapter for feature extraction, and an Auxiliary Feature Extractors (AFE) for calculating additional losses, resulting finally in a mAP@50 of 90.2\% and a mAP of 68.2\% on the HRSID dataset. Under the same context, the present research is situated regarding the existing literature in order to present the contribution of deformable attention mechanisms to the service of a complex task, that of object detection. The basic model, Deformable-DETR \cite{zhu2020deformable}, is proposed to measure its performance against the best-known models in this field, while using two different datasets. Furthermore, this work differs from previous research in that the models presented are applied to both modes of acquisition, relating to optical and SAR images. Moreover, a robust benchmarking protocol is employed, together with the complete code of the benchmarking, enabling users to visualize and compare the results.
\section{Work Benchmarks: Models and Dataset}
\label{main}
In this section, all the models used for the benchmarking pipeline are presented, including those based on classical CNNs and Transformers, together with the used datasets, while introducing the data preparation stage.  
\subsection{Benchmarking models}
A bibliographic study is conducted to explore the models most widely used in previous research, and a customized query is carried out using Scopus platform. Several distinct models emerged from this study, in addition to Deformable-DETR, especially RetinaNet, Faster R-CNN, YOLOv11, as well as transformer-based architectures, which are indeed DETR, DN-DETR, Conditional DETR and DAB-DETR.
\subsubsection{Deformable DETR}
This is a transformer-based model, implemented to solve the DETR \cite{carion2020end} convergence speed problem. As its name suggests, Deformable-DETR \cite{zhu2020deformable} is based on deformable attention mechanism, a modified version of the standard attention, while allowing us to focus only on the most interesting data. On the other hand, a multi-scale attention is employed at various stages of the process (Figure 1), consisting in applying attentions simultaneously on several scales of the input images, while allowing to capture local and global relationships between the extracted features. The attentions in question are calculated according to the following equation: 
\begin{equation}
\text{Attention}(q) = \sum_{m=1}^{M} \sum_{n=1}^{N} A_{mn} \cdot W_{mn} \cdot f(x_{mn}),
\end{equation}
where:
\begin{itemize}
 \item $q$ is the query,
 \item $M$: number of scale levels for feature maps,
 \item $N$: number of sample points,
 \item $A_{mk}$: attention weight, learned by the point n at the level m,
 \item $W_{mn}$: linear projection of the level m and the point n,
 \item $f(x_{mn})$: the feature map in the point $x_{mn}$.
\end{itemize}
\begin{figure}[ht]
  \centering
  \includegraphics[width=0.8\linewidth]{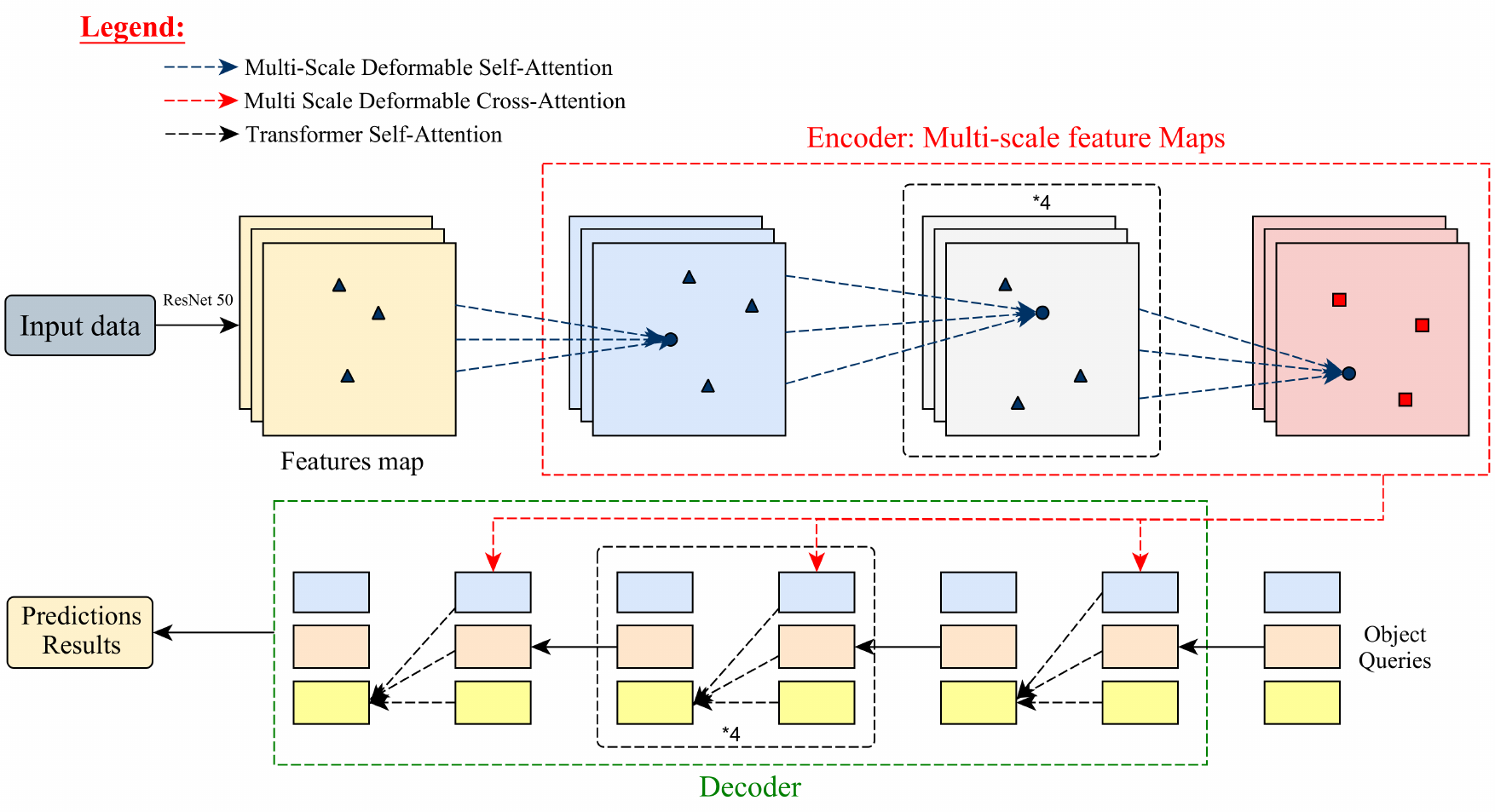} 
  \caption{Deformable-DETR architecture (\cite{zhu2020deformable}).}
  \label{fig:Year_chart}
\end{figure}
\subsubsection{RetinaNet}
Developed by Facebook AI Research (FAIR) in 2017 \cite{lin2017focal}, RetinaNet is a CNN-based detector. First, a ResNet-50 \cite{he2016deep} backbone is deployed for feature extraction, then, a Feature Pyramid Network (FPN) is utilized to create a pyramid of features based on different object scales, the goal behind introducing FPN is to be able to detect objects at different scales. Next, anchor boxes are generated for each level of the pyramid, and a classification and subnet regression are created to concretize the final predictions, while applying the focal loss function to manage the imbalance between positive and negative classes.
\subsubsection{Faster R-CNN}
Presented by Ren et al. \cite{ren2015faster}, Faster R-CNN model also uses a ResNet-50 backbone for feature extraction from training images. Subsequently, a two-stage process is initiated, the first consisting in training a Region Proposal Network (RPN), in order to identify the regions most likely to contain objects, and the second involving the use of RPN outputs to predict final results, precisely classes, and boxes coordinates.
\subsubsection{YOLOv11}
Proposed by Khanam and Hussain \cite{khanam2024yolov11}, YOLOv11 model is the latest version in the You Only Look Once (YOLO) series of detectors. In addition to integrating a customized backbone for feature extraction, by introducing a 3-block convolutional layers with a 2*2 size kernel (C3K2), this new version comprises the implementation of two innovative modules, especially the Spatial Pyramid Pooling - Fast (SPPF) to extract features at different scales and the convolutional block with Parallel Spatial Attention (C2PSA), while introducing spatial attentions in order to guide and optimize training. These innovative approaches have resulted in a more robust accuracy for YOLOv11, when compared with previous versions.
\subsubsection{DEtection TRansformer (DETR)}
Based on Transformers \cite{vaswani2017attention} and introduced by Facebook Artificial Intelligence Research \cite{carion2020end}, DETR is an end-to-end model using an encoder-decoder architecture. A ResNet-50 backbone is firstly used to extract features from input images, followed by a Transformer Encoder to retrieve the global relationships between these features, as well as a Transformer decoder to process the output representations of the encoder. Then, a Feed Forward Network (FFN) is employed to generate the final outputs, i.e the predicted classes and bounding boxes.
\subsubsection{DeNoising DETR (DN-DETR)}
Introduced by Li et al. \cite{li2022dn}, DN-DETR presents an extension of DETR model by adding \textbf{denoising} queries at the level of the decoder architecture. This approach enabled a relative improvement in the performance of DETR, while making the model learn the worst cases of the input data. The model response to denoising queries is characterized by a loss function similar to that applied to normal queries.
\subsubsection{Conditional DETR}
With the aim of accelerating the convergence of DETR, Conditional DETR \cite{meng2021conditional} proposes a novel attention mechanism, while using, at the decoder level, a definition of crossed spatial queries based on antecedent representations. This logic enables the training process to be guided by a cross-attention mechanism, resulting in a relative reduction in execution and inference times.
\subsubsection{Dynamic Anchor Boxes DETR (DAB-DETR)}
It is an innovative architecture proposed by Liu et al. \cite{liu2022dab}, using the principle of query dynamicity. In fact, these queries, as coordinate prediction boxes, are refined as the training progresses. An anchor map is constructed at each layer of the decoder in order to adjust the new prediction values, ie. the \textbf{dynamic anchor boxes}. This concept helps to guide the training process, accelerates overall convergence and increases results accuracy, thanks to the dynamic nature of queries and the progressive adjustment of predictions.
\subsection{Datasets used}
Two datasets are exploited to implement this benchmark, specifically Aircraft Pleiades dataset \cite{airbus2021} and Ship SAR Detection Dataset (SSDD) \cite{zhang2021sar}, for the detection of aircrafts and ships on optical/SAR images. Two object classes are built from the images of each dataset, especially the main object class, i.e. aircraft or ship, as well another class reflecting the truncated object on the images, so as to better reflect the realistic use cases.
\begin{figure}[ht]
  \centering
  \includegraphics[width=0.76\linewidth]{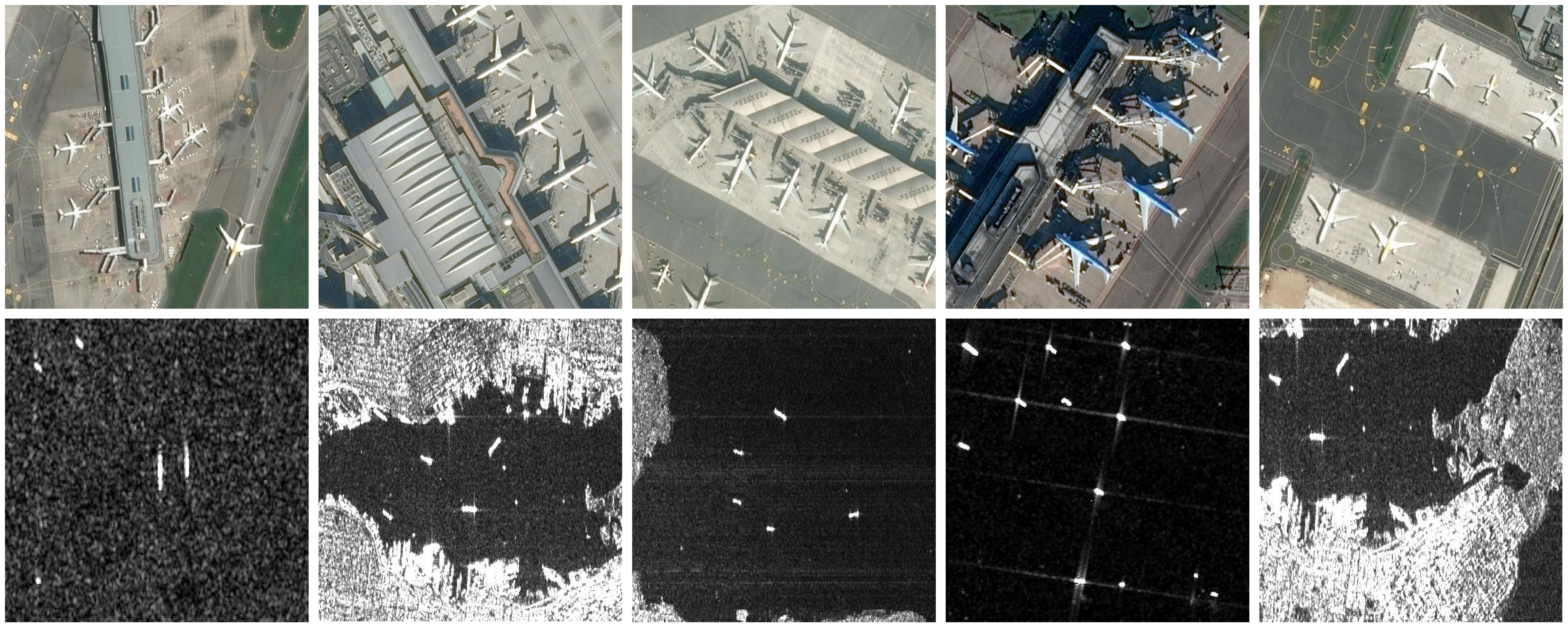} 
  \caption{Sample images of the two datasets (top: Pleiades Aircraft dataset, bottom: SSDD).}
  \label{fig:Year_chart}
\end{figure}
\\
It is important to highlight that an adjustment of the annotations was made, while manually correcting the bounding boxes related to "Truncated Aircraft” and “Truncated Ship” object classes, separating them from the original classes, which was indeed the major problem encountered during this stage. Table 1 shows the characteristics of each dataset, including the resolution of images and the number of instances after correction. This is followed by a phase of automatic data preparation, which consists of standardizing the annotations according to the Common Objects in Context (COCO) format \cite{lin2014microsoft}, a particular annotation standard used so as to train and evaluate various computer vision tasks.
\begin{table}[h]
\caption{Datasets characteristics.}
\begin{tabular*}{\hsize}{@{\extracolsep{\fill}} l l l }
\toprule
\textbf{Dataset} & \textbf{Aircraft Pleiades Dataset} & \textbf{Ship SAR Detection Dataset} \\
\midrule
\textbf{Satellite source} & Pleiades A \& B & RadarSat 2, TerraSARX, Sentinel 1 \\
\textbf{Images Resolution (m)} & 0.5 & From 1 to 15 \\
\textbf{Images Number} & 103 & 1106 \\
\textbf{Classes (Annotations Number)} & \begin{tabular}[c]{@{}l@{}}Airplane (276) \\ Truncated Airplane (14) \end{tabular} & 
\begin{tabular}[c]{@{}l@{}}Ship (2303) \\ Truncated Ship (153) \end{tabular} \\
\bottomrule
\end{tabular*}
\end{table}
\section{Methodology}
In the following section, the approach adopted is presented in detail, as well as the metrics used during the test phase, together with the reasons for their choice. Figure 3 shows the benchmarking steps, including training, validation and test stages.
\begin{figure}[ht]
  \centering
  \includegraphics[width=\linewidth]{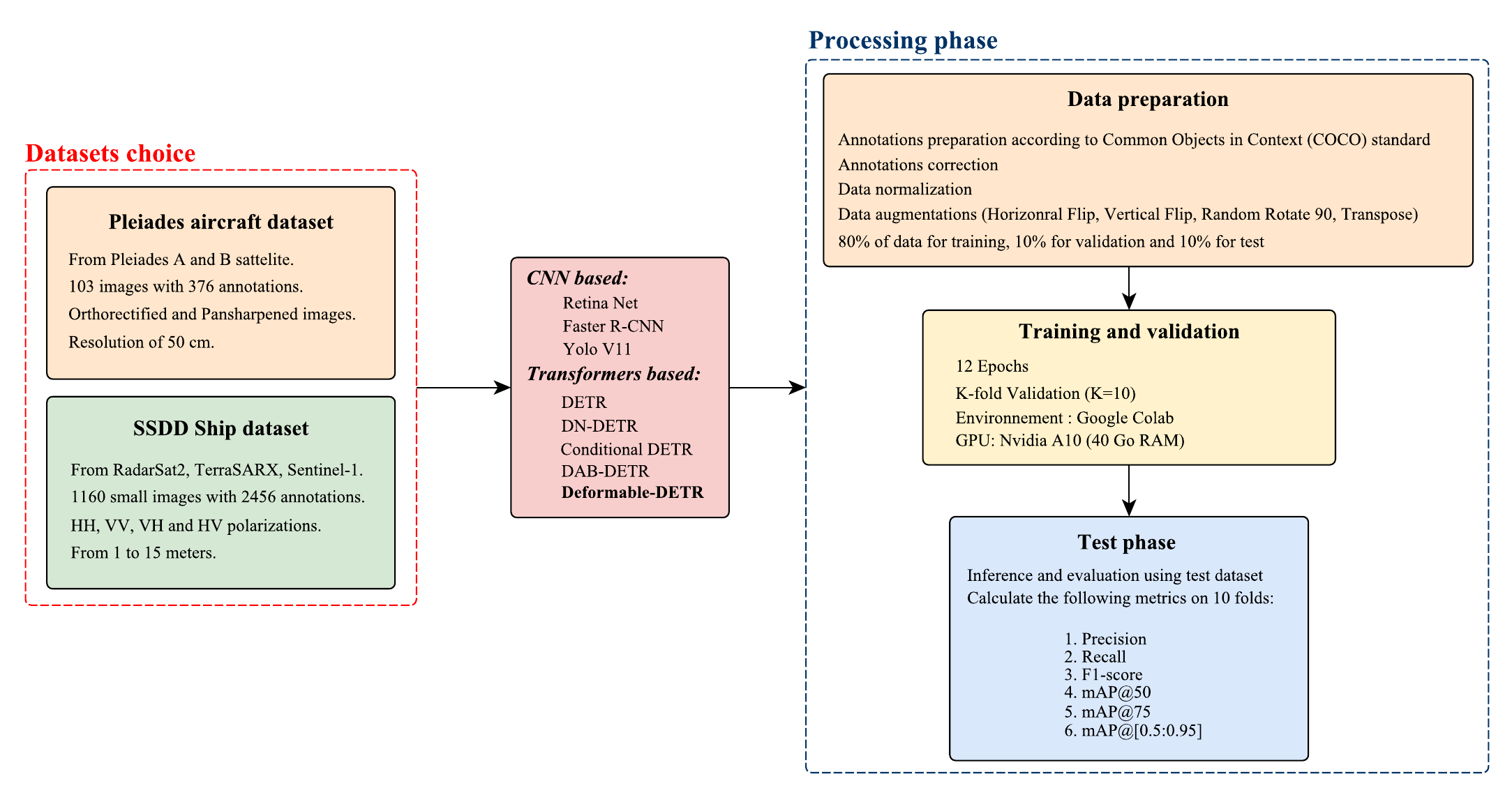} 
  \caption{Methodology workflow.}
  \label{fig:Year_chart}
\end{figure}
As part of data augmentations, transformations are performed on the images/annotations of each dataset during the training and validation. The aim behind these transformations, which are horizontal flip, grayscale, i.e. converting a color image from the RGB mode to grayscale, and Gaussian Blur, which involves the application of image blurring based on a Gaussian distribution, is to improve the robustness of produced models while following a more realistic scenario. Moreover, pre-trained models on COCO dataset \cite{lin2014microsoft} are exploited for the fine-tuning, while adjusting the relative weights of all pre-trained model layers, starting from the initial weights.
\subsection{Used metrics}
For the test phase, the following metrics are employed:
\paragraph{\underline{\textbf{Precision}}}
This is a measure quantifying the proportion of correct positive predictions among the set, i.e.: 
\begin{equation}
    \text{Precision} = \frac{TP}{TP + FP},
\end{equation}
where TP and FP are the proportions of true positives and false positives, respectively.
\paragraph{\underline{\textbf{Recall}}}
Unlike precision, recall measures the model's ability to detect all true positives. In other words, this metric calculates the proportion of true positives in relation to the total number of predictions:
\begin{equation}
    \text{Recall} = \frac{TP}{TP + FN}.
\end{equation}
It is noted that FN corresponds to false negatives results.
\paragraph{\underline{\textbf{F1-Score}}}
It is a harmonic mean of precision and recall, based on the formula:
\begin{equation}
    F1\text{-}Score = 2 \times \frac{\text{Precision} \times \text{Recall}}{\text{Precision} + \text{Recall}}.
\end{equation}
The F1-Score is very useful when class proportions are unbalanced, as it represents a strict balance between precision and recall, hence its choice for our case.
\paragraph{\textbf{\underline{mean Average Precision (mAP) metrics}}}
mAP corresponds to the Average Precisions (APs) for the dataset classes on different IOUs. For N classes, the mAP is:
\begin{equation}
    mAP = \frac{1}{N} \cdot \sum_{i=1}^{N} AP_i.
\end{equation}
Note that AP is the area under the precision–recall curve, for a given IoU threshold:
\begin{equation}
    AP = \int_{0}^{1} \text{Precision}(Recall) \, dRecall.
\end{equation}
Therefore, three metrics are used to compare the performance of models, especially:
\begin{itemize}
    \item mAP@50, i.e. the average mAP for IoU $\geq$ 0.5,
    \item mAP@75, corresponding to mAPs values for IoU $\geq$ 0.75,
    \item mAP@[0.5:0.95], noted simply mAP, corresponding to the average of mAPs between 0.5 and 0.95 IoU values.
\end{itemize}
\section{Experimental results}
80\% of the images are used for model training, over exactly 12 epochs, 10\% are used for validation, and the same for the test phase. To scan the inferred models over the entire dataset, a stratified 10-Fold validation is performed, consisting in choosing a balanced number of instances for each iteration, with respect to object classes. Test metrics are then calculated for each fold, and averaged to obtain a more global estimate of performance. It is noted that the whole process is run using an NVIDIA A100 GPU with 40 GB RAM.
\\
Test results (Table 2) proved Deformable-DETR model to be the best performer. Over 12 epochs only, the involved model is trained over 365 seconds for the optical dataset and 3750 seconds for the SSDD. Deformable-DETR achieved F1-Scores of 95.12\% and 94.54\%, and mAP metrics of 76.75\% and 76.14\% on Pleiades and SSD datasets, respectively.
\begin{table}[h]
\caption{Test results for datasets used (in \%).}
\small                                    
\begin{tabular*}{\textwidth}{@{\extracolsep{\fill}} 
  >{\raggedright\arraybackslash}p{1cm}  
  >{\raggedright\arraybackslash}p{2.25cm}  
  >{\centering\arraybackslash}p{1cm}    
  >{\centering\arraybackslash}p{1cm}    
  >{\centering\arraybackslash}p{1cm}    
  >{\centering\arraybackslash}p{1cm}      
  >{\centering\arraybackslash}p{1cm}      
  >{\centering\arraybackslash}p{1cm}      
  >{\centering\arraybackslash}p{1cm}      
  >{\centering\arraybackslash}p{2cm}}   
\toprule
Dataset & Metric & RetinaNet & Faster R-CNN & YOLOv11 & DETR & DN-DETR & Conditional DETR & DAB-DETR & \textbf{Deformable DETR} \\
\midrule
\multirow{7}{*}{\shortstack{\textbf{Pleiades}\\\textbf{Aircraft}}}
  & Training Time (s) & 719.46 & 435.22 & 342.39 & 327.34 & 354.15 & 338.93 & 329.36 & \textbf{306.53} \\
  & Precision         & 80.80  & 93.33  & 96.01  & 93.21  & 92.54 & 94.01 & 95.23 & \textbf{97.76}  \\
  & Recall            & 73.3   & 93.33  & 92.55  & 90.35  & 87.6 & 88.48 & 90.06 & \textbf{92.62}  \\
  & F1-Score          & 76.87  & 93.33  & 94.27  & 91.19  & 89.9 & 91.1 & 92.57 & \textbf{95.12}  \\
  & mAP@50            & 74.76  & 81.66  & 98.16  & 80.05  & 83.16 & 83.84 & 95.14 & \textbf{98.42}  \\
  & mAP@75            & 69.56  & 78.97  & 76.33     & 78.46  & 80.14 & 82.26 & 87.53 & \textbf{89.42}  \\
  & mAP               & 55.80  & 73.77  & 66.18  & 73.44  & 73.81 & 73.98 & 74.19 & \textbf{76.75}  \\
\midrule
\multirow{7}{*}{\shortstack{\textbf{SSDD}\\\textbf{SAR}}} 
  & Training Time (s) & 3997.13 & 3452.22 & 3401.79 & 3583.12 & 3834.89 & 3576.29 & 3489.27 & \textbf{3370.16} \\
  & Precision         & 83.85   & 90.36   & 94.29   & 86.31   & 90.54 & 94.01 & 95.23 & \textbf{96.26}   \\
  & Recall            & 83.13   & 87.23   & 92.14   & 89.21   & 91.36 & 89.19 & 92.49 & \textbf{92.88}   \\
  & F1-Score          & 82.80   & 88.77   & 93.18   & 87.74   & 90.95 & 91.54 & 93.84 & \textbf{94.54}   \\
  & mAP@50            & 86.75   & 91.61   & 96.81   & 95.42   & 92.18 & 94.63 & 96.14 & \textbf{97.31}   \\
  & mAP@75            & 78.02   & 73.93   & 83.25   & 84.12   & 84.14 & 86.78 & 87.89 & \textbf{88.66}   \\
  & mAP               & 64.54   & 62.53   & 76.03   & 73.86   & 73.18 & 74.89 & 74.54 & \textbf{76.14}   \\
\bottomrule
\end{tabular*}
\normalsize                                   
\end{table}
\begin{figure}[ht]
  \centering
  \includegraphics[width=0.9\linewidth]{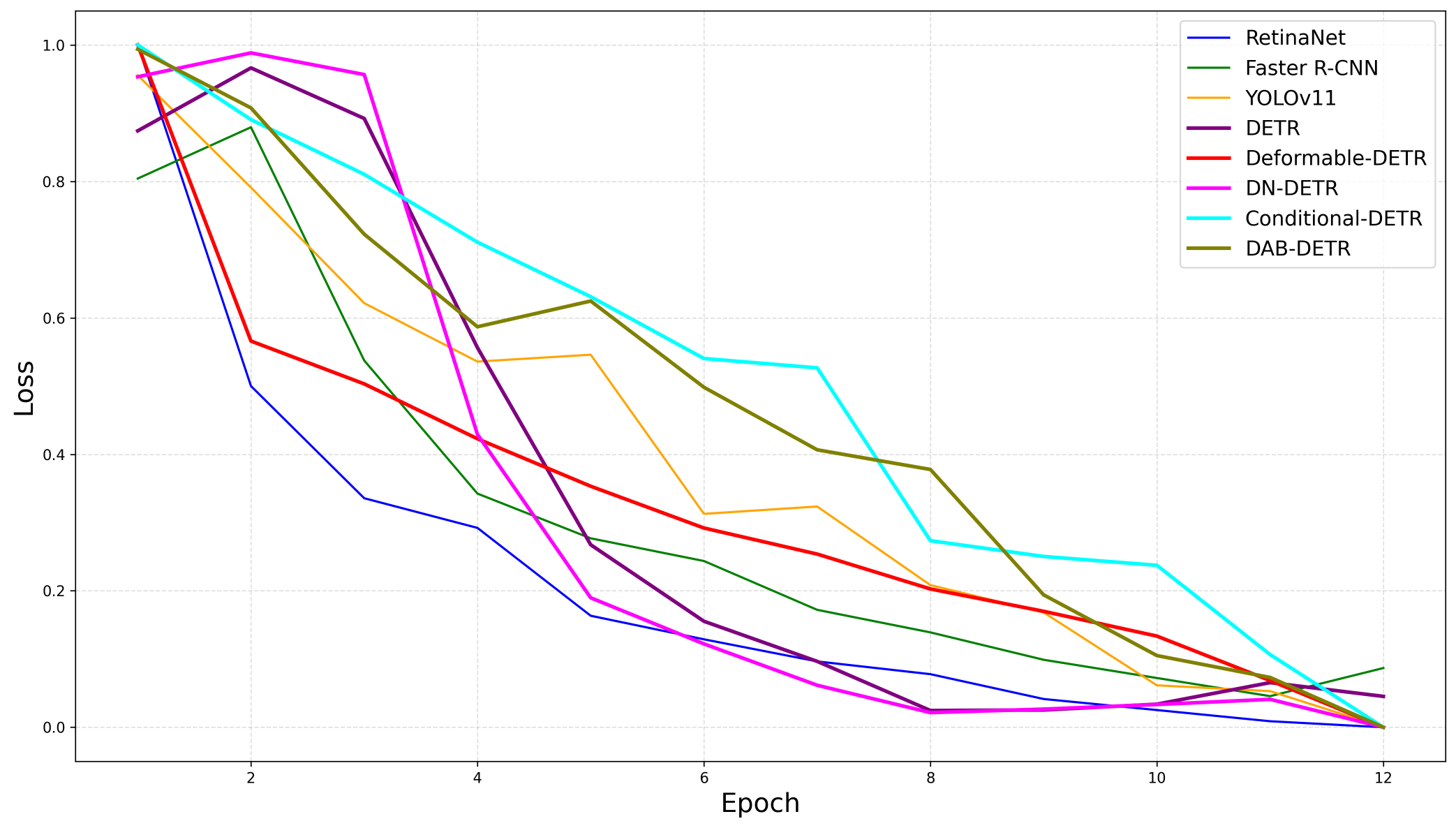} 
  \caption{Combined Normalized Loss (Regression + Classification) across models.}
  \label{fig:Year_chart}
\end{figure}
\noindent 
\\
Figure 4 shows the combined losses for each epoch, obtained by normalizing the values for each model and averaging the losses over the two datasets. It can be seen from this graph that the loss for Deformable-DETR seems to be more stable than for the others, since the associated curve is strictly decreasing. Also, it is observed that the convergence of Deformable-DETR is much faster, given the slope of the associated loss curve. Additionally, Figure 5 illustrates examples of detection results (bounding boxes in green), compared with input images and bounding boxes of the ground truth (in yellow). As these examples show, Deformable-DETR has a notable detection capability, particularly in the case of small-scale objects. The multi-scale deformable attention distinguishing our proposition is well demonstrated by the presented inference examples.
\\
It is noted that the datasets, as well as the complete code for this benchmark, in the form of training and inference notebooks for each model, are available on this Github repository\footnote{\url{https://github.com/BOUTAYEBAnasse/RS-Deformable-DETR-Benchmarking}}.
\begin{figure}[ht]
  \centering
  \includegraphics[width=\linewidth]{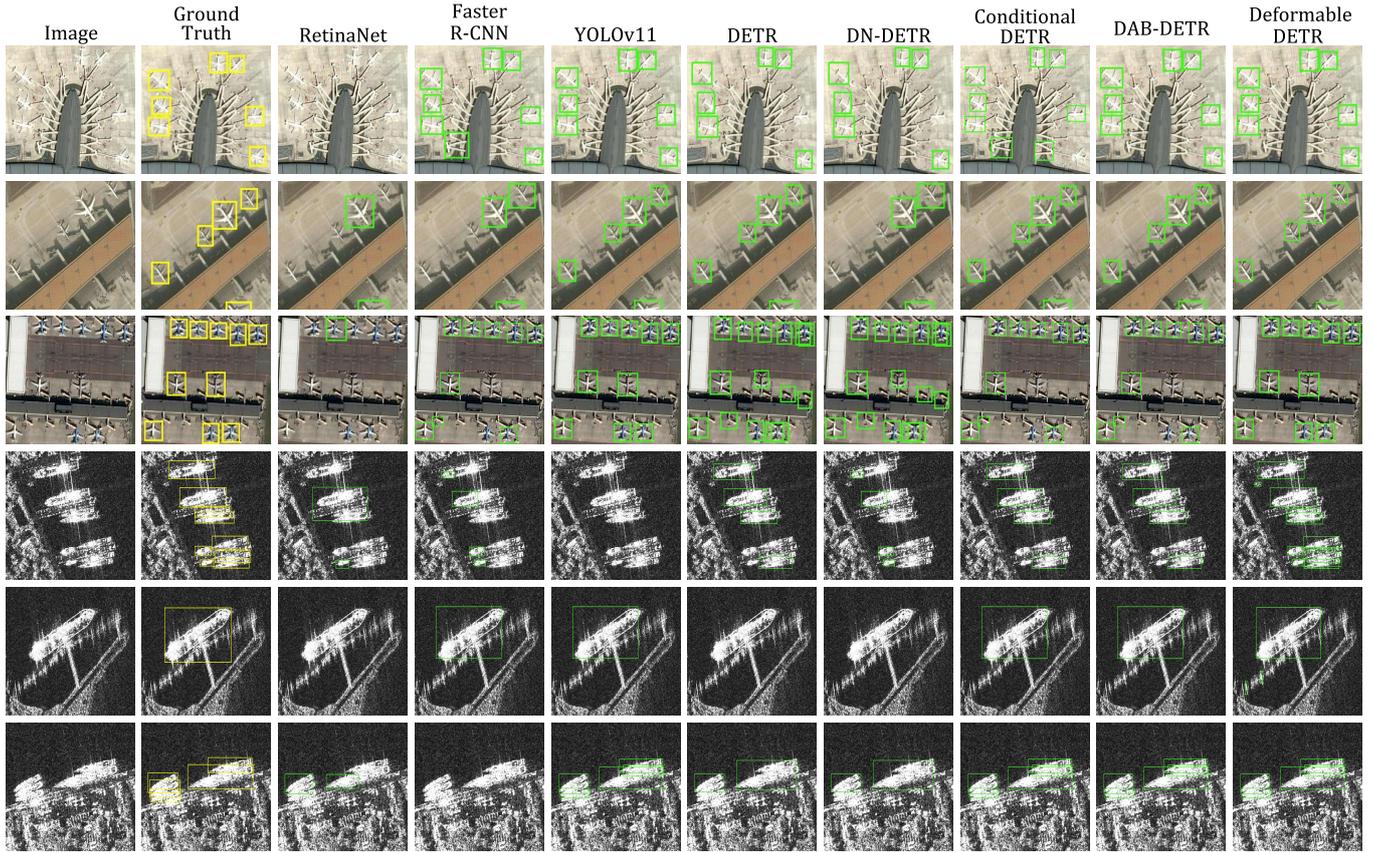} 
  \caption{Detection results on the two datasets.}
  \label{fig:Year_chart}
\end{figure}
\\
Moreover, a specific performance validation was executed while comparing the Deformable DETR model with other transformer-based models, but this time dedicated specifically to the detection task on remotely sensed images, namely ASAFE \cite{shi2022adaptive}, Dense Attention Pyramid Network (DAPN) \cite{cui2019dense} and CRTTransSar \cite{shi2022local}. To comply with the benchmarking protocol employed for the above-mentioned models, a fine-tuning is implemented, using pre-trained models on ImageNET-1K dataset \cite{deng2009imagenet}, while starting the adjustment from the backbone layers (ResNet-50) over 200 epochs. Table 3 shows the related benchmarking results on the original SSD dataset, i.e. without annotations correction, which proves the value of the Deformable-DETR model, even in the face of more specialized models.
\begin{table}[h]
\caption{Comparison of object detection models on remote sensing images (metrics are in \%)}
\begin{tabular*}{13cm}{@{\extracolsep{\fill}}%
                     p{3.5cm}   
                     p{1.8cm}   
                     p{1.8cm}   
                     p{1.8cm}   
                     p{1.8cm}}  
\toprule
\textbf{Model} & \textbf{mAP@50} & \textbf{Precision} & \textbf{Recall} & \textbf{F1-Score} \\
\midrule
ASAFE            & 95.19 & 88.54 & 95.94 & 92.09 \\
DAPN             & 90.60 & 85.60 & 91.40 & 88.40 \\
CRTTransSar      & 94.70 & 92.13 & 90.46 & 91.29 \\
\textbf{Deformable-DETR} & \textbf{98.13} & \textbf{97.55} & \textbf{95.19} & \textbf{96.36} \\
\bottomrule
\end{tabular*}
\end{table}

\FloatBarrier 
\section{Conclusion and future work}
This paper evaluates the performance of deformable attention mechanisms, through Deformable-DETR model, while comparing it with the best-performing models for the task of object detection, whether based on CNNs or Transformers, or conceptualized specifically for remote sensing. Experimental results from a K-fold validation proved the particular performance of the model in question, in terms of detection quality, based on evaluation metrics, and in terms of training speed. It is therefore concluded that deformable attention mechanisms, through the notions of multi-scale attention and guided scores calculation, are perfectly suited to target detection problems on remotely sensed images, particularly with regard to the differences in object scales, as well as the optimization of computational costs.
\\
As a matter of fact, an interesting prospect would be the implementation of a similar benchmark for the image segmentation task, which is useful for situating the best-performing models in question. Furthermore, using transformer-based classifiers, with robust backend for richer feature extraction, would be future work that would prove beneficial. In this context, several transformer-based architectures have proven their performance against CNN-based ones, such as Data-efficient Image Transformer (DeiT) \cite{touvron2021training} or Swin Transformer \cite{liu2021swin}.

\bibliography{references}

\end{document}